\documentclass[10pt,twocolumn,letterpaper]{article}

\usepackage{cvpr}
\usepackage{times}
\usepackage{epsfig}
\usepackage{graphicx}
\usepackage{color}
\usepackage{amsmath}
\usepackage{amssymb}
\usepackage[ruled,vlined,boxed]{algorithm2e}
\usepackage{multirow}

\usepackage[pagebackref=true,breaklinks=true,letterpaper=true,colorlinks,bookmarks=false]{hyperref}

\cvprfinalcopy 

\def\cvprPaperID{1468} 

\ifcvprfinal\fi
\begin{document}

\title{SegNet: A Deep Convolutional Encoder-Decoder Architecture for Robust Semantic Pixel-Wise Labelling}

\author{Vijay Badrinarayanan, Ankur Handa, Roberto Cipolla\\
Machine Intelligence Lab, Department of Engineering,\\
University of Cambridge, UK\\
{\tt\small vb292,ah781,cipolla@eng.cam.ac.uk}
}
\maketitle

\begin{figure*}[!t]
\center
\includegraphics[width=0.95\textwidth]{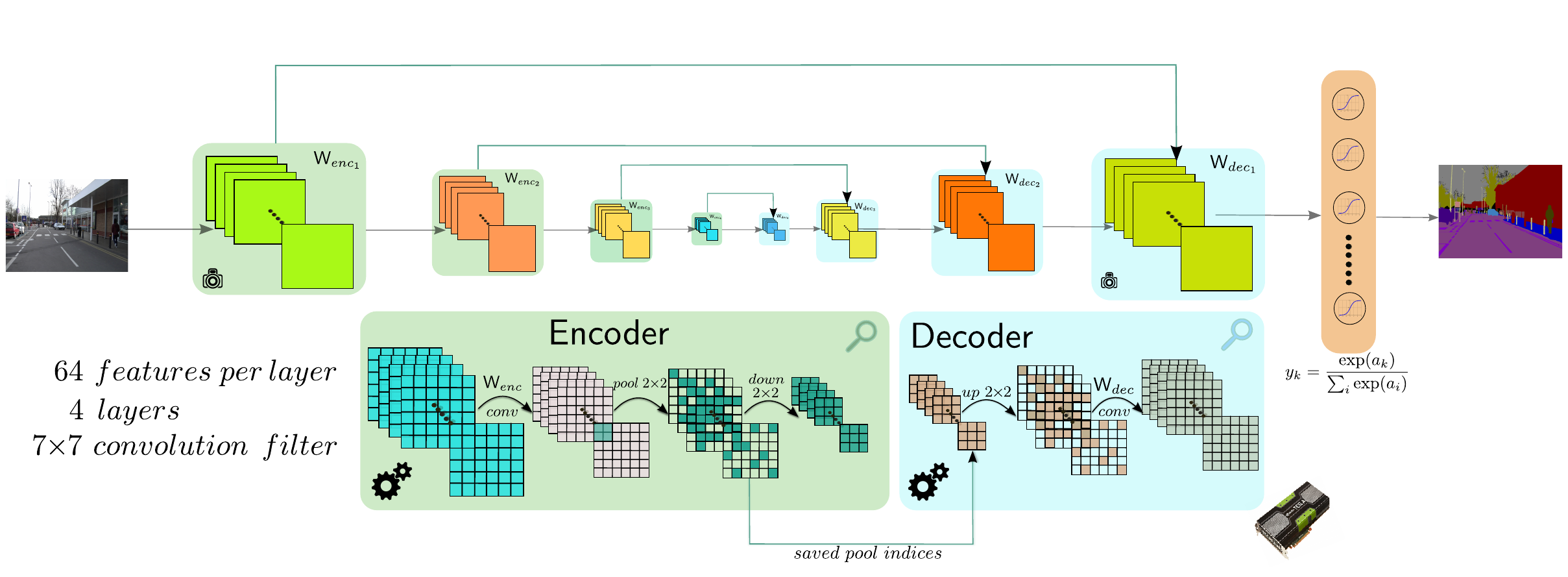}
\caption{\footnotesize{A 4 layer SegNet which takes in an RGB input image and performs \textit{feed-forward} computation to obtain pixel-wise labelling. A stack of feature encoders is followed by a corresponding decoders. The soft-max layer classifies each pixel independently using the features input by the last decoder. An encoder uses the convolution-ReLU-max pooling-subsampling pipeline. A decoder upsamples its input using the transferred pool indices from its encoder. It then performs convolution with a trainable filter bank.}}
\label{SegNetArchitecture}
\end{figure*} 


\begin{abstract}
We propose a novel deep architecture, SegNet, for semantic pixel wise image labelling \footnote{This version was submitted to CVPR' 15 on November 14, 2014 with paper Id 1468. A similar architecture was proposed http://arxiv.org/pdf/1505.04366.pdf on May 17, 2015.}. SegNet has several attractive properties; (i) it only requires forward evaluation of a fully learnt function to obtain smooth label predictions, (ii) with increasing depth, a larger context is considered for pixel labelling which improves accuracy, and (iii) it is easy to visualise the effect of feature activation(s) in the pixel label space at any depth. \\
SegNet is composed of a stack of encoders followed by a corresponding decoder stack which feeds into a soft-max classification layer. The decoders help map low resolution feature maps at the output of the encoder stack to full input image size feature maps. This addresses an important drawback of recent deep learning approaches which have adopted networks designed for object categorization for pixel wise labelling. These methods lack a mechanism to map deep layer feature maps to input dimensions. They resort to ad hoc methods to upsample features, \textit{e.g.} by replication. This results in noisy predictions and also restricts the number of pooling layers in order to avoid too much upsampling and thus reduces spatial context. SegNet overcomes these problems by learning to map encoder outputs to image pixel labels. We test the performance of SegNet on outdoor RGB scenes from CamVid, KITTI and indoor scenes from the NYU dataset. Our results show that SegNet achieves state-of-the-art performance even without use of additional cues such as depth, video frames or post-processing with CRF models. 
\end{abstract}




\vspace{-3mm}
\section{Introduction}
Semantic segmentation is an important step towards understanding and inferring different objects and their arrangements observed in a scene. This has wide array of applications ranging from estimating scene geometry, inferring support-relationships among objects to autonomous vehicle driving. Early methods that relied on low-level vision cues have fast been superseded by popular machine learning algorithms. In particular, deep learning has seen huge success lately in handwritten digit recognition, speech, categorising whole images and detecting objects in images \cite{DBLP:journals/corr/SzegedyLJSRAEVR14, Simonyan} also seen growing interest in semantic pixel-wise labelling problems \cite{FarabetPAMI,raey,socher2011parsing}. However, these recent approaches have tried to directly adopt deep architectures designed for category prediction to pixel-wise labelling. The results, although very encouraging, have not been quite satisfactory. Primarily, the deepest layer representations/feature maps are of a small resolution as compared to input image dimensions due to several pooling layers \textit{e.g.} if $2\times2$ non-overlapping max-pooling-subsampling layers are used three times, the resulting feature map is $1/8^{th}$ of the input dimension. Therefore, an ad hoc technique is used to upsample the deepest layer feature map to match the input image dimensions by replicating features within a block \textit{i.e.} all pixels within a block ($8\times8$ in our example) have the same features. This often results in predictions that appear blocky\footnote{see http://david.grangier.info/scene\_parsing/}. This is exactly what we improve using our proposed SegNet architecture, wherein the decoders learn to map the deepest layer features to full image dimensions. Learning to decode has two other advantages.
First, deeper layers each with pooling-subsampling can be introduced which increases the spatial context for pixel labelling. This results in smooth predictions unlike patch based classifiers \cite{Sturgess, Brostow}. Second, ablation studies to understand the effects of features such as in \cite{Zeiler} can be performed using the decoder stack.

We draw inspiration of our encoder-decoder type architectures from probabilistic auto-encoders used to build generative models \cite{NgiamMultiModal} and unsupervised learning of feature hierarchies \cite{Ranzato}. Our main contribution is to learn an encoder-decoder stack trained in a \textit{modular and fully supervised} manner for pixel-wise labelling. The addition of each deeper encoder-decoder pair results in an increased spatial context \textit{i.e.}, a $4$ layer SegNet with $7\times7$ kernels and $2\times2$ non-overlapping max pooling in each layer has a spatial context of $106\times106$ pixels when a feature-map is backtracked to the input image. The SegNet predictions get smoother as more layers are added and demonstrate high accuracy, comparable to or even exceeding methods which use CRFs \cite{Sturgess}. SegNet maintains a constant number of features per layer which is typically set to $64$. This has a practical advantage that the computational cost successively decreases for each additional/deeper encoder-decoder pair.

In Sec. \ref{LitReview} we review related recent literature. We describe in detail the SegNet architecture in Sec. \ref{Architecture} along with its qualitative analysis. Our quantitative experiments with SegNet on several well known benchmark datasets are described in Sec. \ref{Experiments}. We also discuss the advantages and drawbacks of our approach including computational times. We conclude with pointers to future work in Sec. \ref{Conclusion}. For most of our experiments, we use outdoor RGB road scene analysis \cite{GabeDataset, GeigerKITTI} and indoor RGBD scene analysis \cite{silberman2012indoor} datasets to measure the quantitative performance.
 

 
\section{Literature Review}
\label{LitReview}
Semantic pixel-wise segmentation is an ongoing topic of research, fuelled by challenging datasets \cite{GabeDataset,silberman2012indoor,GeigerKITTI}. Current best performing methods all mostly rely on hand engineered features generally used for per-pixel independent classification. Typically, a patch is fed into a classifier \textit{e.g.} Random Forest \cite{Jamie2,Brostow} or Boosting \cite{Sturgess,LadickyECCV} to predict the class probabilities of the center pixel. Features based on appearance \cite{Jamie2}, SfM and appearance \cite{Brostow,Sturgess, LadickyECCV} have been explored for the CamVid test. These per-pixel noisy predictions (often called \textit{unary} terms) from the classifiers are then smoothed by using a pair-wise or higher order CRF \cite{Sturgess,LadickyECCV} to improve the accuracy. More recent approaches have aimed to produce high quality unaries by trying to predict the labels for all the pixels in a patch as opposed to only the center pixel. This improves the results of Random Forest based unaries \cite{kontschieder2011structured} but thin structured classes are classfied poorly. Dense depth maps computed from the CamVid video have also been used as input for classification using Random Forests \cite{zhang2010semantic}. Another approach argues for the use of a combination of popular hand designed features and spatio temporal super-pixelization to obtain higher accuracy \cite{tighe2013superparsing}. Recent top performing technique on the CamVid test \cite{LadickyECCV} addresses the imbalance among label frequencies by using additional training data from the PASCAL VOC dataset to learn object detectors. The result of all these techniques indicates the need for improved classification as increases in accuracy have mostly come from adding new features or modalities to the classifier. Post-processing using CRF models of various orders \cite{Sturgess} has mainly resulted in improving the accuracy of dominant classes such as sky, road, buildings with little effect on the accuracy of thin structured but equally important classes such as signs, poles, pedestrians. This highlights the need for better pixel-wise classification  when imbalanced label frequencies exist.\\
Meanwhile, indoor RGBD pixel-wise semantic segmentation has also gained popularity since the release of the NYU dataset \cite{silberman2012indoor} which showed the usefulness of the depth channel to improve segmentation. Their approach used features such as RGB-SIFT, depth-SIFT, location as input to a neural network classifier to predict pixel unaries. The noisy unaries are then smoothed using a CRF. Improvements were made using a richer feature set including LBP and region segmentation to obtain higher accuracy \cite{ren2012rgb} followed by a CRF. In more recent work \cite{silberman2012indoor}, both class segmentation and support relationships are inferred together using a combination of RGB and depth based cues. Another approach focusses on real-time joint reconstruction and semantic segmentation, where Random Forests are used as the classifier \cite{Hermans14ICRA}. Gupta et al. \cite{gupta2013perceptual} use boundary detection and hierarchical grouping before performing category segmentation. The common attribute along all these approaches is the use of hand engineered features for pixel-wise classifiction of either RGB or RGBD images. \\
The application of deep learning for scene segmentation has only just begun. There have also been a few attempts to apply networks designed for categorization to segmentation, particularly by replicating the deepest layer features in blocks to match image dimensions \cite{FarabetPAMI,FarabetPurityCover,Grangier,Gatta}. However, the resulting classification is blocky \cite{Grangier}. Another approach using recurrent neural networks \cite{pinheiro2014recurrent} merges several low resolution predictions to create input image resolution predictions. On the whole, although some of these techniques already present improvements over hand engineered features \cite{FarabetPAMI}.

Our work is inspired by the unsupervised feature learning architecture proposed by Ranzato \textit{et. al} \cite{Ranzato}. The key learning module is an encoder-decoder network where the encoder consists of a filter bank convolution, tanh squashing function, max pooling followed by sub-sampling to obtain the feature maps. For each sample, the indices of the max locations computed during pooling are stored and passed to the decoder. The decoder upsamples the feature maps by using the already stored pooled indices, also called switches, and learns a decoder filter bank to reconstruct the input image. This architecture was used for unsupervised pre-training of feature hierarchies. A similar decoding technique is used for visualizing trained convolutional networks\cite{zeiler2010deconvolutional} for object classification; the transposed encoder kernels are set as the decoder kernels which are followed by a non-linearity and the pooling indices are used for upsampling. The architecture of Ranzato mainly concentrated on layer wise feature learning using small input patches although during test time a full sized image was the input. This discrepancy was corrected for by Kavukcuoglu et. al. \cite{KorayUnsup} by using test size images/feature maps to learn hierarchical encoders. Both these approaches however did not attempt to use \textit{deep encoder-decoder} networks for unsupervised feature training as they discarded the decoders after each encoder training. Here, the SegNet architecture differs from these approaches as the objective used for training all the encoder-decoder pairs is the same, \textit{i.e.}, to minimise the cross-entropy label loss.

Other applications where pixel wise predictions are made using deep networks are image super-resolution \cite{dong2014learning} and depth map prediction from a single image \cite{eigen2014depth}. The authors in \cite{eigen2014depth} discuss the need for learning to upsample from low resolution feature maps which is the central topic of this paper.

\begin{figure*}
\centering
\includegraphics[width=0.9\textwidth]{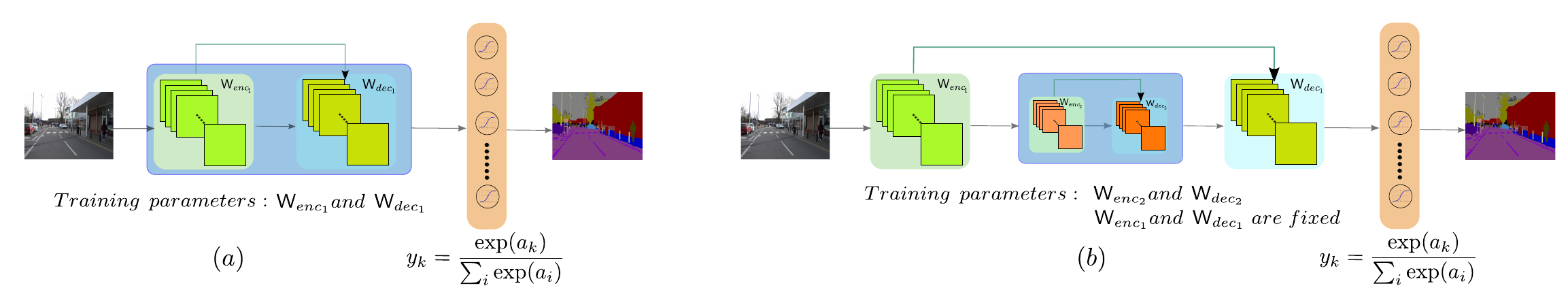}
\caption{\footnotesize{(a) The modular training starts by optimizing the first encoder and decoder weights. The soft-max can be pre-trained or randomly initialized. (b) Once the first pair is trained, we \textit{insert} an inner deeper encoder-decoder pair and optimize these weights while holding the outer encoder-decoder and soft-max weights fixed. More deeper pairs are then subsequently trained. Note encoder and decoder weights are untied.}}
\label{SegNetTrain}
\end{figure*} 



\section{SegNet Architecture and Learning Scheme}
\label{Architecture}
A four layer SegNet architecture used in our experiments is illustrated in Fig. \ref{SegNetArchitecture}. Each encoder performs dense convolutions, ReLU non-linearity, a non-overlapping max pooling with a $2\times2$ window and finally down-sampling. Each decoder upsamples its input using the memorized pooled indices and convolves it  with a trainable filter bank. No ReLU non-linearity is used in the decoder unlike the deconvolution network \cite{Zeiler,zeiler2010deconvolutional}. This makes it easier to optimize the filters in each pair. The encoder and decoder filters are also untied to provide additional degrees of freedom to minimize the objective. The final layer is a soft-max classifier (with no bias term) which classifies each pixel independently. The output of the soft-max is a K channel image where K is the number of classes. 

SegNet uses a ``flat" architecture, \textit{i.e}, the number of features in each layer remains the same ($64$ in our case) but with full connectivity. This choice is motivated by two reasons. First, it avoids parameter explosion, unlike an expanding deep encoder network with full feature connectivity (same for decoder). Second, the training time remains the same (in our experiments it slightly decreases) for each additional/deeper encoder-decoder pair as the feature map resolution is smaller which makes convolutions faster. Note that the decoder corresponding to the first encoder (closest to the input image) produces a multi-channel feature map although the encoder input is either 3 or 4 channels (RGB or RGBD) (see Fig. \ref{SegNetArchitecture}). This high dimensional feature representation is fed to the soft-max classifier. This is unlike the other decoders which produce feature maps the same size as their encoder inputs. A fixed pooling window of $2\times2$ with a stride of non-overlapping $2$ pixels is used. This small size preserves thin structures in the scene. Further, a constant kernel size of $7\times7$ over all the layers was chosen to provide a wide context for smooth labelling \textit{i.e.} a pixel in the deepest layer feature map can be traced back to a context window in the input image of $106\times106$  pixels. The trade-off here is between the size of the context window and retaining thin structures. Smaller kernels decrease context and larger ones potentially destroy thin structures.

The input to the SegNet can be any arbitrary multi-channel image or feature map(s), \textit{e.g.}, RGB, RGBD, map of normals, depth etc. We perform local contrast normalization (LCN) as a pre-processing step to the input \cite{Simoncelli,Jarrett}. The advantage of this step are many, (i) to correct for non-uniform scene illumination thus reducing the dynamic range (increases contrast in shadowed parts). (ii) highlighting edges which leads the network to learn category shape, (iii) improves convergence as it decorrelates the input dimensions \cite{Simoncelli}. LCN is performed independently for each modality, \textit{i.e.}, RGB is contrast normalized as a three channel input and depth as a single channel for RGBD inputs. This avoids highlighting pseudo depth edges due to RGB edges and vice-versa. 

\subsection{Training the SegNet}
\label{Training}
Most deep learning methods use stochastic gradient descent (SGD) for training \cite{lecun1998gradient}. SGD needs sufficient expertise to initialize weights with appropriate magnitudes, adapting appropriately learning rates and momentum parameters which both control the step sizes. Therefore, we adopt L-BFGS \cite{Nocedal} based on the comparative study by Ngiam \textit{et. al} \cite{LeDeepOptim} who advocate the use of L-BFGS particularly for auto-encoders. L-BFGS has faster and more stable convergence than SGD. It also works well in large batches which is useful to maximize the throughput of powerful GPUs. We initialize the weights in all the layers and the soft-max weights from a zero mean unit variance Gaussian $\mathcal{N}(0,1)$ and normalized the kernels to unit L2 norm. We obtained good predictive performance from the network without the need for special layer-wise weight initialization or any learning rate tuning. We also use inverse frequency weighting for the classes to correct for any label imbalances in the training set \cite{Jamie2}. 

We use mini-batches that maximize GPU usage and avoid GPU-CPU memory transfers. Typically, $25-50$ randomly chosen images (with replacement) per mini-batch. The optimizer is run for $20$ iterations per mini-batch and $10$ epochs for each layer. We empirically observe that the objective plateaus after $5-6$ epochs and so we run another $4$ epochs as a margin. Note that, after $10$ epochs, each input sample approximately ``influences" the optimizer 
$200$ times.  \\
We train the encoder-decoder pair weights closest to the input layer. The soft-max layer can be trained first or randomly initialised. It then remains fixed throughout the experiment. Next, we introduce a deeper layer of encoder-decoder (see Fig. \ref{SegNetTrain}) and train their weights while holding the shallower layer encoder-decoder weights fixed. Note that the objective remains the same, \textit{i.e.}, to minimize label cross-entropy loss over the mini-batch. This is unlike unsupervised feature learning approaches which reconstruct the input of the layer in question \cite{Ranzato,KorayUnsup}, thus varying the objective with each layer. The deconvolution network \cite{zeiler2010deconvolutional} on the other hand optimizes the same reconstruction objective with each deeper layer. The difference to our approach is (i) the objective is unsupervised, (ii) there is no encoder to learn a feed-forward representation thus requiring an optimisation step during test time to produce features for recognition. We successively add deeper encoder-decoder pairs and train them while holding the preceeding pair's weights fixed. In total, we use 4 layer networks, \textit{i.e.}, 4 encoders and 4 decoders in our experiments. Once the encoder-decoder stack is trained, we find that there is no advantage to training the soft-max layer as it only relies on a linear discriminant function.\\
We wrote our own Matlab GPU compatible implementation of SegNet that uses the minFunc optimization library \cite{minFunc}. Our code has been tested on NVIDIA Tesla K40, GTX GeForce 880M and GTXGeForce780 GPUs. We will make our light-weight Matlab code available publicly soon. With the current state of code optimisation, training a 4 layer deep SegNet on the CamVid dataset (367 training images of $360\times480$) takes about a week. The unoptimized test time is in the order of $2$secs/frame: bulk of the computation time is spent performing tensor convolutions in the feedforward path and FFT based convolutions during backpropagation \footnote{more speedup can be gained \url{https://developer.nvidia.com/cuDNN}}. 

\begin{figure*}
\centering
\includegraphics[width=0.9\textwidth]{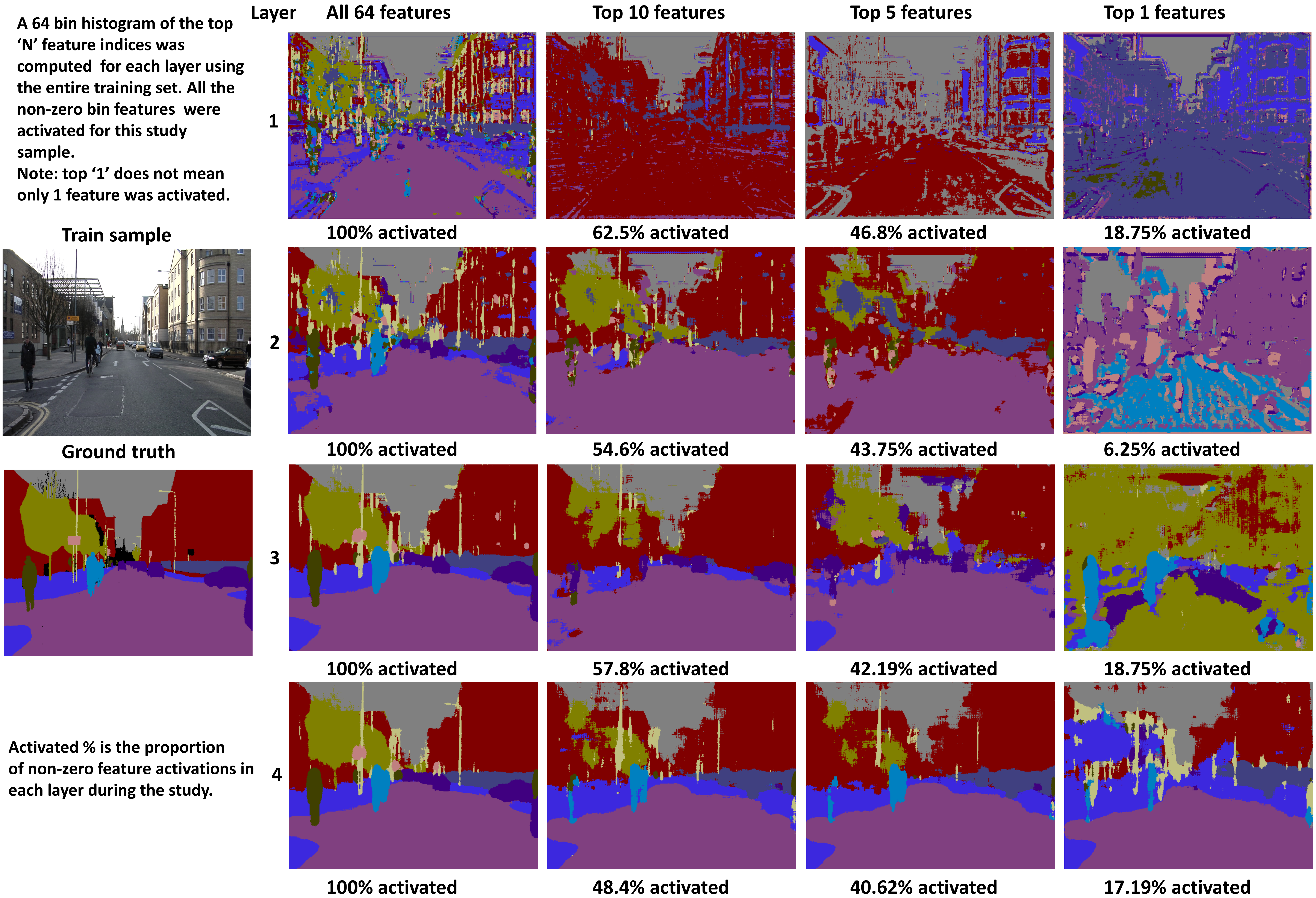}
\caption{\footnotesize{SegNet feature ablation study. All layers use 64 features in the SegNet. The four columns from right to left show the predictions obtained at various depths when only a fraction of the feature activations are used and the remaining set to zero. Note the quality of the labelling improves strikingly at depth 4 even when only the top-1 feature activations are considered. Interestingly, these activations seem to be tuned largely for the static scene classes and the other classes such as cars are labelled only when more features are activated. When fewer features are activated, missing categories (cars) are filled with sidewalk which is reasonable. Also shown are the percentage of activated features as part of the top 'N' activations; deeper layers have fewer but more finely tuned activations.}}
\label{SegNetAblation}
\end{figure*}

\subsection{Visualizing the SegNet}
\label{Visualization}
We perform an ablation study to gain some insight into about the SegNet features. The work of Zeiler et al. \cite{Zeiler} study the effects of feature activations in each layer of a trained network \cite{krizhevsky2012imagenet}. The feature activations are mapped back to image pixel space using a deconvolutional network. The SegNet architecture by construction is trained to decode the encoder activations and we use this to visualize the effect of feature activations (which layer) in the pixel label space. A recent study \cite{szegedy2013intriguing}  has shown that in each layer of a deep network it is the ``direction" or ``space" (ensemble of feature activations) which encodes useful class information rather than individual units (feature activations). We therefore focus our study on the predictive effect of a subset of feature activations at each layer. \\
For a given layer, we compute the feature activations/maps for each sample in the training set. We then compute the root mean square value of each map \textit{i.e.} $\forall j \in\{1..64\} $ $\sqrt{ \frac{1}{N}\sum_{i\in \mathcal{I}} (f_{j}^{i})^{2} }$ where $f_j^i$ is $j^{th}$ feature map value at pixel $i$ at a given layer. This assigns each map a single value \textit{e.g.}, the CamVid training set would have a $64$ dimensional vector for each training sample for layer 4 of the SegNet. We now compute a histogram of the top `N' elements of each such vector over all the samples. This histogram shows the most activated features in that layer over the training set. For any `N',  we set the remainder of feature maps to zero (ablation) and decode the pixel-wise labelling for a given input sample. Note that since our training is modular, this can be done after each deeper layer has been added. Some results of the top 'N' feature activations based labelling across all the layers are shown in Fig. \ref{SegNetAblation}. \\
We observe firstly that the predictions get smoother as depth is increased which is a consequence of larger spatial context in the input space. More interestingly, the top-1 4th layer features predict almost entirely the static scene classes and ``fill in" the missing cars \textit{e.g.} with sidewalk. Given the feature(s) which get activated for cars are zeroed out, this prediction is reasonable and indicates the network is able to learn spatial context/class location information. Similarly, trees are \textit{filled in} with buildings and bollards are extended to poles. In contrast, this effect is less clear and gets worse for shallower layers. This suggests subsets of features in the deeper layers are more ``tuned" to certain scene categories in agreement with earlier work \cite{Zeiler}.\\
We would like to add here that our efforts to perform an ablation study by choosing each feature map in turn and setting the remaining to zero produced results which were not clearly interpretable. It is also interesting to note that for shallower layers to produce qualitatively better predictions 'N' has to be set to about 5 or 10. The corresponding histogram has atleast $50\%$ of the features activated as opposed to about $15\%$ for the top-1 in layer 4, indicating deeper features are tuned to groups of related categories.

\begin{table*}[tch]

\small{

\begin{tabular}{|l|l|l|l|l|l|l|l|l|l|l|l|l|l|}

\hline

Method                   & \rotatebox{90}{Building} & \rotatebox{90}{Tree} & \rotatebox{90}{Sky}  & \rotatebox{90}{Car}  & \rotatebox{90}{Sign-Symbol} & \rotatebox{90}{Road} & \rotatebox{90}{Pedestrian} & \rotatebox{90}{Fence} & \rotatebox{90}{Column-Pole} & \rotatebox{90}{Sidewalk} & \rotatebox{90}{Bicyclist} & \rotatebox{90}{Class avg.} & \rotatebox{90}{Global avg.} \\ \hline


SfM+Appearance  \cite{Brostow}           & 46.2     & 61.9 & 89.7 & 68.6 & 42.9        & 89.5 & 53.6       & 46.6  & 0.7         & 60.5     & 22.5      & 53.0       & 69.1        \\ \hline

Boosting    \cite{Sturgess}              & 61.9     & 67.3 & 91.1 & 71.1 & \textbf{58.5}        & 92.9 & 49.5       & 37.6  & 25.8        & 77.8     & 24.7      & 59.8       & 76.4        \\ \hline

Dense Depth Maps   \cite{zhang2010semantic}          & 85.3     & 57.3 & 95.4 & 69.2 & 46.5        & \textbf{98.5} & 23.8       & 44.3  & 22.0        & 38.1     & 28.7      & 55.4       & 82.1        \\ \hline

Structured Random Forests \cite{kontschieder2011structured}& \multicolumn{11}{l|}{not available}                                                                          & 51.4       & 72.5        \\ \hline

Neural Decision Forests \cite{BuloNeural}  & \multicolumn{11}{l|}{not available}                                                                          & 56.1       & 82.1        \\ \hline

Local Label Descriptors  \cite{yang2012local}  & 80.7     & 61.5 & 88.8 & 16.4 & n/a         & 98.0 & 1.09       & 0.05  & 4.13        & 12.4     & 0.07      & 36.3       & 73.6        \\ \hline

Super Parsing   \cite{tighe2013superparsing}           & \textbf{87.0}     & 67.1 & 96.9 & 62.7 & 30.1        & 95.9 & 14.7       & 17.9  & 1.7         & 70.0     & 19.4      & 51.2       & 83.3        \\ \hline

%
%

SegNet - 4 layer          & 75.0     & \textbf{84.6} & 91.2 & \textbf{82.7} & 36.9        & 93.3 & \textbf{55.0}       & 37.5  & \textbf{44.8}        & 74.1     & 16.0      & \textbf{62.9}       & \textbf{84.3}        \\ \hline


Boosting + pairwise CRF  \cite{Sturgess} & 70.7     & 70.8 & 94.7 & 74.4 & 55.9        & 94.1 & 45.7       & 37.2  & 13.0        & 79.3     & 23.1      & 59.9       & 79.8        \\ \hline

Boosting+Higher order \cite{Sturgess}    & 84.5     & 72.6 & \textbf{97.5} & 72.7 & 34.1        & 95.3 & 34.2       & 45.7  & 8.1         & 77.6     & 28.5      & 59.2       & 83.8        \\ \hline

Boosting+Detectors+CRF \cite{LadickyECCV}   & 81.5     & 76.6 & 96.2 & 78.7 & 40.2        & 93.9 & 43.0       & \textbf{47.6}  & 14.3        & \textbf{81.5}     & \textbf{33.9}      & 62.5       & 83.8        \\ \hline

\end{tabular}

}
\caption{\footnotesize{Quantitative results on CamVid \cite{GabeDataset}. We consider SegNet - 4 layer for comparisons with other methods. SegNet performs best on several challenging classes (cars, pedestrians, poles) while maintaining competitive accuracy on remaining classes. The class average and global average is the highest even when compared to methods using structure from motion \cite{Brostow}, CRF \cite{Sturgess,LadickyECCV}, dense depth maps \cite{zhang2010semantic}, temporal cues \cite{tighe2013superparsing}.}}
\label{CamVidQuant}
\end{table*}

\begin{table*}[tch]

\small{

\begin{tabular}{|l|l|l|l|l|l|l|l|l|l|l|l|l|l|l|l|}

\hline

Method              & \rotatebox{90}{Bed}  & \rotatebox{90}{Objects} & \rotatebox{90}{Chair} & \rotatebox{90}{Furniture} & \rotatebox{90}{Ceiling} & \rotatebox{90}{Floor} & \rotatebox{90}{Decoration} & \rotatebox{90}{Sofa} & \rotatebox{90}{Table} & \rotatebox{90}{Wall} & \rotatebox{90}{Window} & \rotatebox{90}{Books} & \rotatebox{90}{TV}   & \rotatebox{90}{Class avg.} & \rotatebox{90}{Global avg.} \\ \hline

Multi-scale Convnet \cite{FarabetPurityCover} & 38.1 & 8.7     & 34.1  & \textbf{42.4}      & 62.6    & 87.3  & 40.4       & 24.6 & 10.2  & \textbf{86.1} & 15.9   & 13.7  & 6.0  & 36.2       & 52.4        \\ \hline

3D Semanic Mapping \cite{Hermans14ICRA} & \textbf{68.4} & 8.6     & \textbf{41.9}  & 37.1      & \textbf{83.4}    & \textbf{91.5}  & 35.8       & \textbf{28.5} & \textbf{27.7}  & 71.8 & \textbf{46.1}   & \textbf{45.4}  & \textbf{38.4} & \textbf{48.0}       & \textbf{54.2}        \\ \hline
%
%
%

SegNet - layer 4    & 40.5 & \textbf{27.7}    & 39.9  & 36.1      & 70.5    & 82.4  & \textbf{55.0}       & 22.7 & 24.9  & 61.5 & 38.6   & 15.2  & 17.9 & 41.0       & 50.5             \\ \hline

\end{tabular}

}
\caption{\footnotesize{Quantitative results on the NYU v2 \cite{silberman2012indoor}. The SegNet performs better than the multi-scale convnet which uses the same inputs (and post-processing) on 9 out of 13 classes. The method in \cite{Hermans14ICRA} uses additional cues such as ground plane detect and column-wise depth normalization to achieve better accuracy.}}.
\label{NYUQuant}
\end{table*}

\begin{table}[h]
\footnotesize{
\begin{tabular}{|l|l@{\hskip -0.005mm}|l@{\hskip -0.005mm}|l@{\hskip -0.005mm}|l@{\hskip -0.005mm}|l@{\hskip -0.005mm}|l@{\hskip -0.005mm}|l@{\hskip -0.005mm}|l@{\hskip -0.005mm}|l@{\hskip -0.005mm}|}

\hline

Method     & \rotatebox{90}{Building} & \rotatebox{90}{Tree} & \rotatebox{90}{Sign} & \rotatebox{90}{Road} & \rotatebox{90}{Fence} & \rotatebox{90}{Pole} & \rotatebox{90}{Sidewalk} & \rotatebox{90}{Class avg.} & \rotatebox{90}{Global avg.} \\ \hline

Space-Time CRF \cite{SpanishKITTI} &\hspace{-2mm}  84.3 & \hspace{-2mm} \textbf{92.8} & \hspace{-2mm} 17.1 & \hspace{-2mm} 96.8 & \hspace{-2mm} \textbf{62.9}  & \hspace{-2mm} 2.1  & \hspace{-2mm} \textbf{75.2}     & \hspace{-2mm} \textbf{61.6}       & \hspace{-2mm} 85.6        \\ \hline

SegNet(R)                     & \hspace{-2mm} \textbf{88.6} \hspace{-2mm}     & \hspace{-2mm} 92.2 & \hspace{-2mm} \textbf{23.3} & \hspace{-2mm} \textbf{97.4} & \hspace{-2mm} 9.2   & \hspace{-2mm} 39.9 & \hspace{-2mm} 69.4     & \hspace{-2mm} 60.0       & \hspace{-2mm} \textbf{89.7}        \\ \hline

SegNet(L4)                       & \hspace{-2mm} 74.1     & \hspace{-2mm} 84.9 & \hspace{-2mm} 12.3 & \hspace{-2mm} 93.4 & \hspace{-2mm} 17.0   & \hspace{-2mm} \textbf{56.8} & \hspace{-2mm} 69.4     &  \hspace{-2mm} 58.4      & \hspace{-2mm} 83.0        \\ \hline

SegNet (SM) & \hspace{-2mm} 46.1 & \hspace{-2mm} 79.7 & \hspace{-2mm} 16.3 &\hspace{-2mm}  58.9 & \hspace{-2mm} 30.4 & \hspace{-2mm} 50.1 & \hspace{-2mm} 24.9 & \hspace{-2mm} 43.8 & \hspace{-2mm} 58.7 \\ \hline

\end{tabular}
}
\caption{\footnotesize{Quantitative results on the KITTI dataset \cite{GeigerKITTI,SpanishKITTI}. The SegNet performance is better globally and comparable among classes. The fence class resembles buildings and needs other cues such as temporal information used in \cite{SpanishKITTI} for better accuracy.}}
\label{KITTIQuant}
\end{table}

\begin{figure*}
\centering
\includegraphics[width=0.9\textwidth]{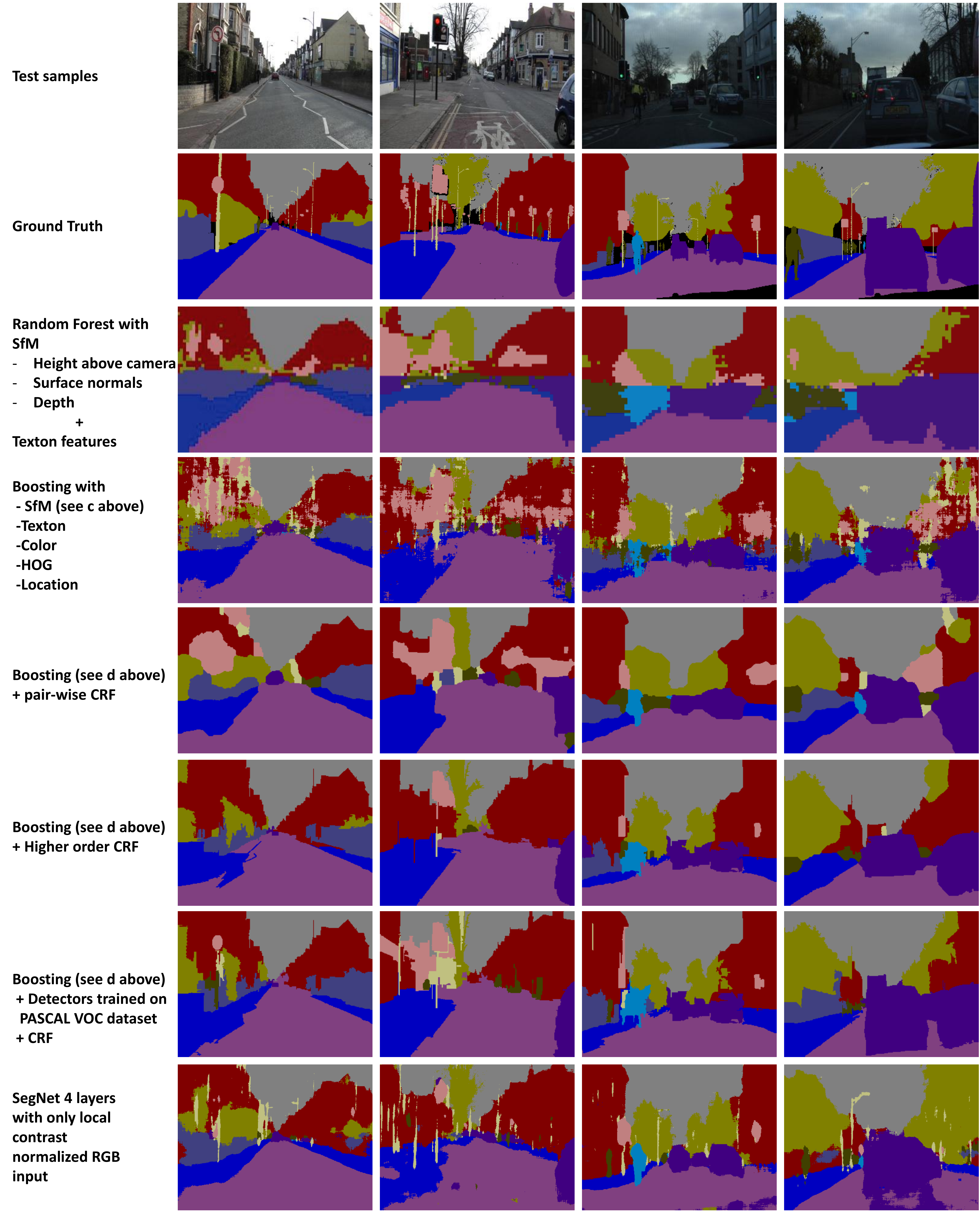}
\caption{\footnotesize{Result samples on CamVid day and dusk test sequences. The evolution of various unary predictions and unaries combined with externally trained detectors \cite{LadickyECCV} and CRF models \cite{Sturgess}. SegNet predictions retain small categories such as poles (column 2,4),  bicyclist (column 3), far side sidewalk (column 2) better than other methods while producing overall smooth predictions. CRF results, although smooth, miss several important categories even when SfM based cues are used. In the dusk scenario, SfM cues are particularly valuable (row 3). Here the SegNet fails to label the car (column 4) however, it fills this part with very reasonable predictions.}}
\label{CamVidQualy}
\end{figure*}

\begin{figure}
\centering
\includegraphics[width=0.5\textwidth]{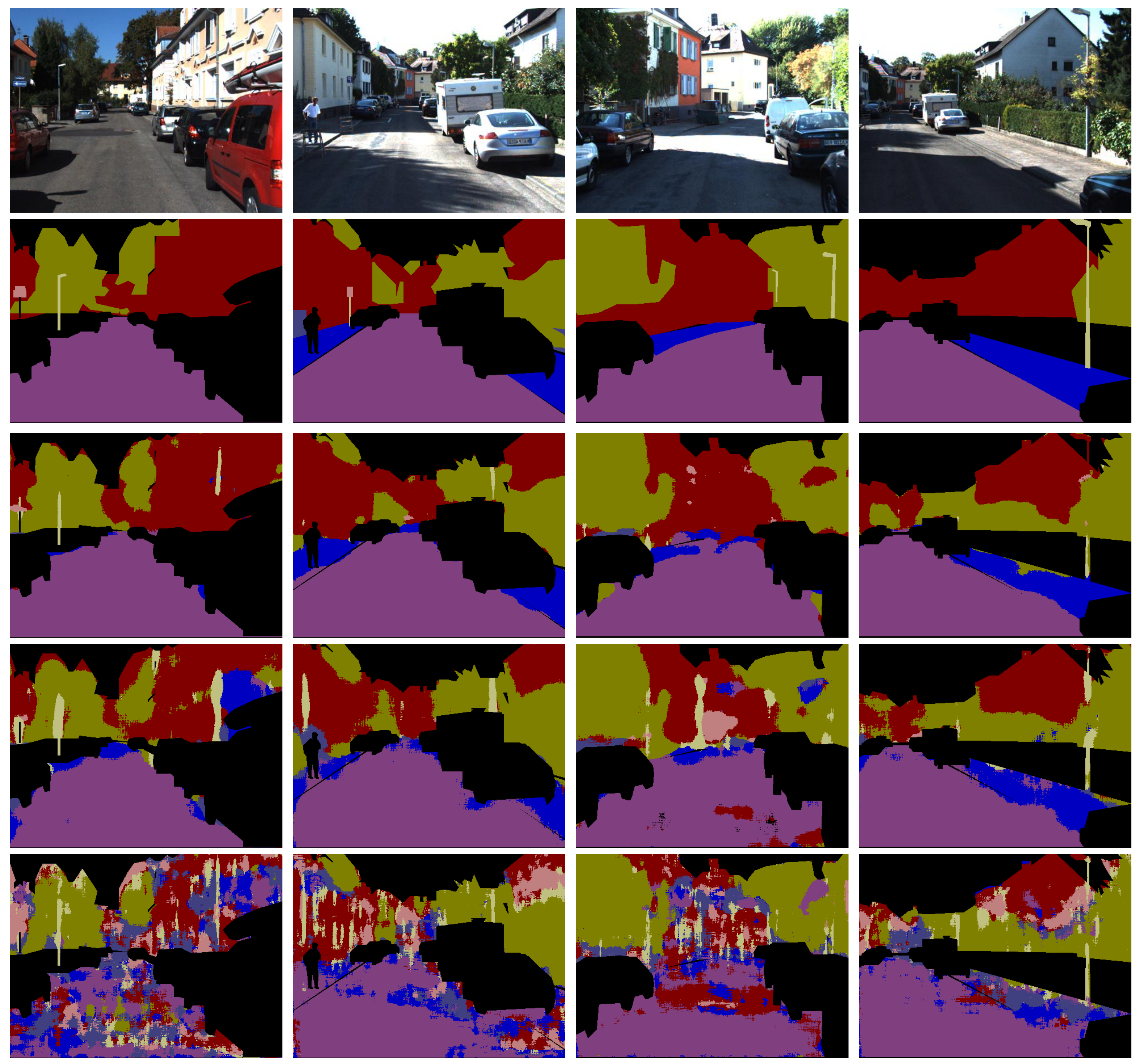}
\caption{\footnotesize{Row 1,2 show KITTI test samples. Notice the illumination differences to CamVid samples in Fig. \ref{CamVidQualy}. Row 3: Predictions when all layers are trained afresh from a random initialization using the KITTI training set. Row 4: SegNet is pre-trained with the CamVid dataset and only layer 4 is trained for two epochs on the KITTI training set. Supervised pre-training can produce good results with a small extra computational effort. Row 5: poor results obtained when starting from pre-trained weights and training only a soft-max classifier with a hidden layer. Unknown class is blackened.}}
\label{KITTIQualy}
\end{figure}

\begin{figure}
\centering
\includegraphics[width=0.5\textwidth]{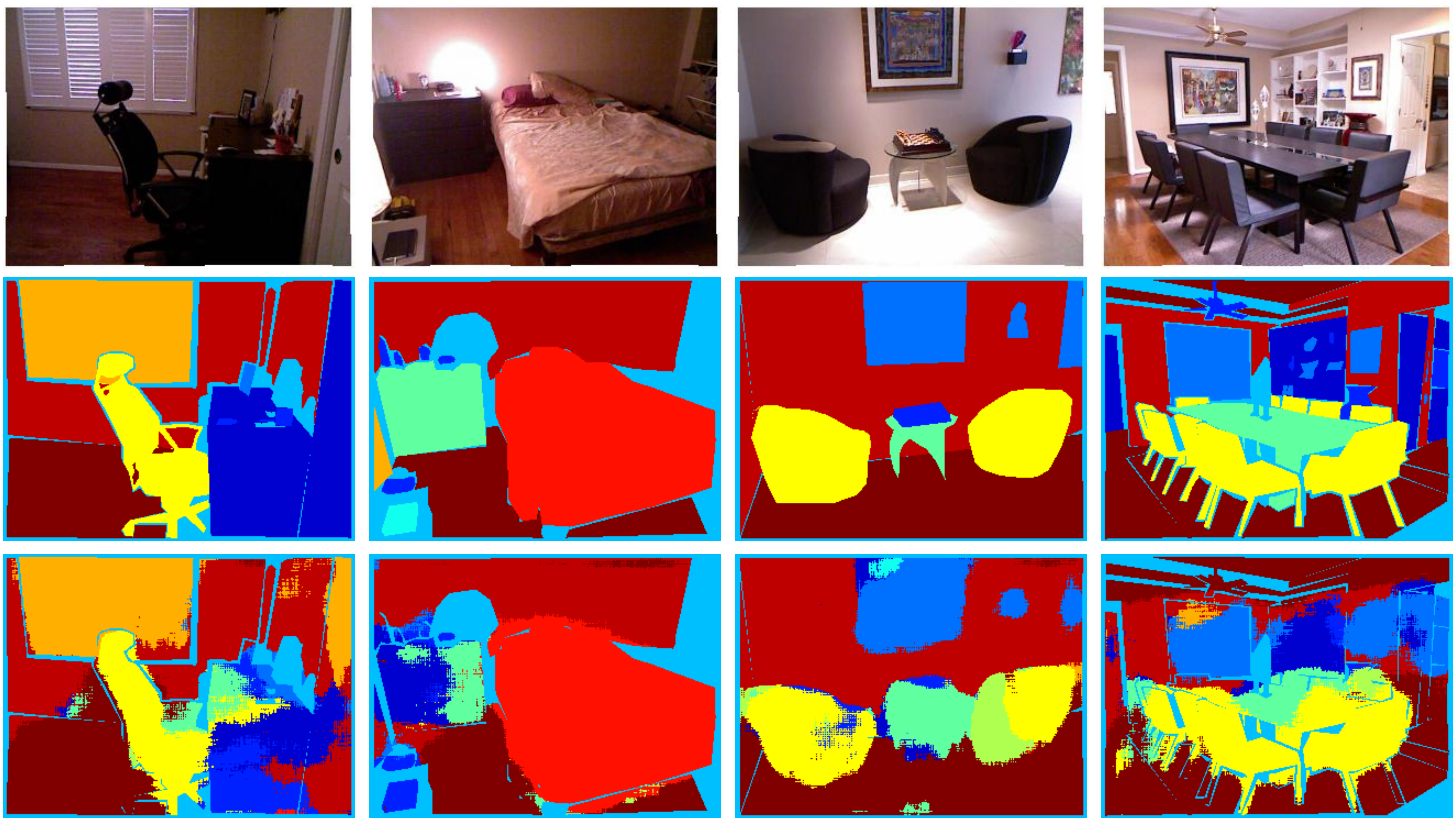}
\caption{\footnotesize{Row 1,2: Indoor scenes from NYU dataset version 2 and their ground truth. Row 3: SegNet predictions with RGBD input. No extra information such as ground plane fitting, column wise pixel depth normalization \cite{Hermans14ICRA} or multi-scale inputs \cite{FarabetPurityCover} was used. Although the predictions are largely correct, the edges between categories are not sharp mainly because of low input resolution, interpolated depth values close to class edges.}}
\label{NYUQualy}
\end{figure}

\section{Experiments and Analysis}
\label{Experiments}
A number of outdoor scene datasets are available for semantic parsing \cite{gould2009decomposing,russell2008labelme,GabeDataset,GeigerKITTI}. Out of these, we chose the CamVid \cite{GabeDataset} and KITTI \cite{GeigerKITTI} datasets which contains 11 semantic classes such as road, building, cars, pedestrians etc.. There is a large imbalance in their frequencies \cite{GabeDataset}. Road, Sky, Building pixels are approximately $40-50$ times more than pedestrian, poles, sign-symbols, cars, bicyclists in the dataset making it very challenging to label smaller categories. This dataset contains video sequences, thus we are able to benchmark our approach with those which use motion and structure \cite{LadickyECCV,Sturgess,Brostow} and video segments \cite{tighe2013superparsing}. Other datasets have more balanced label frequencies and are still image datasets. Another reason for choosing CamVid as compared to SIFT-flow, LabelMe is that the size of the training set is small ($367$) making it feasible to train the SegNet given a standard GPU in reasonable time. The CamVid dataset also contains train and test images ($233$) in day and dusk (poor lighting) conditions. The qualitative comparisons of SegNet predictions with several well known algorithms (unaries, unaries+CRF) are shown in Fig. \ref{CamVidQualy}. The qualitative results show the ability of the SegNet to segment small (cars, pedestrians, bicyclist) classes while producing a smooth segmentation of the overall scene. The other methods shown in Fig. \ref{CamVidQualy} use structure from motion based cues. Lacking this cue, the SegNet misses some labels (cars) but fills it in with other reasonable context related classes. The CRF based results are smooth but do not retain small classes. More dense models \cite{koltun2011efficient} can be better but with additional cost of inference. Table \ref{CamVidQuant} compares the algorithms numerically and demonstrates its superiority over recent competing methods. 
\\
The KITTI dataset is the largest publicly available road scene dataset. Recently, some images from this dataset have been hand-labelled ($8$ classes) for inferring dense 3D semantic maps \cite{SpanishKITTI}. Note that the image sizes are approximately, $376\times1241$, and so we cropped the centre $360\times480$ to make it compatible with the CamVid dataset. We use this dataset to analyse the effect of supervised pre-training using the CamVid data on the KITTI test set. First, we add here that testing on the KITTI samples with only the pre-trained SegNet (using CamVid data) resulted in poor performance. This is because of illumination related differences between the datasets. Therefore, we experimented with three other training variants for the KITTI dataset; (i) training all the layers of the SegNet from a random initialization, denoted SegNet(R), (ii) initializing the parameters with CamVid trained values and training only a soft-max classifier with a hidden layer, denoted SegNet(SM), and (iii) initializing the parameters with CamVid trained values and training only the 4th layer of the SegNet for just $2$ epochs, denoted SegNet(L4). High quality predictions are obtained in scenario SegNet(R) as expected (Fig. \ref{KITTIQualy}). The good performance with CamVid pre-training and layer 4 training shows that, (i) useful semantic cues can be transferred across datasets using the shallower layers, and (ii) it is beneficial to train the deepest layer of the SegNet first given a small computational budget. Table \ref{KITTIQuant} shows the SegNet(R) is competitive even when temporal cues \cite{SpanishKITTI} are not used.\\
For indoor RGBD scenes, the NYU dataset (version 2)\cite{silberman2012indoor} is the largest benchmark dataset containing $795$ training and $654$ testing images with $14$ class (objects, furniture, wall, ceiling etc.) labelling comparison. The NYU dataset has been used to benchmark Farabet et. al's \cite{FarabetPAMI} multi-scale deep learning approach to scene parsing. This benchmark is therefore useful to compare their method, which uses ad hoc feature upsampling, with our learning to upsample based approach. We also note that they learn approximately $1.2M$ parameters as compared to SegNet's $1.4M$ parameters. Other methods either use the smaller NYU dataset \cite{ren2012rgb}, different performance measures \cite{gupta2013perceptual} or test on a small set of classes cite{raey}. The quantitative analysis shown in Table \ref{NYUQuant} show that the SegNet predictions are better the multi-scale convnet (2 pooling layers only) in 9 out of 13 classes. This suggests the SegNet can deal with scale changes by increasing context using deeper layers. The overall results are still far from satisfactory and the lack of cues such as height from ground, depth normalization (used in \cite{Hermans14ICRA}) are needed to achieve better performance. The qualitative results in Fig. \ref{NYUQualy} show that the predictions are largely correct but lack sharp edges. This is due to low input resolution of $320\times240$, lack of ground truth around class edges,and errors in depth interpolation. Another reason is that over the different datasets we tested on, the parameters of the SegNet remained the same. We plan to study the NYU dataset in more detail in the future. Additional results can be viewed in the supplementary material.

\section{Conclusion}
\label{Conclusion}
We presented SegNet, a fully trainable deep architecture for joint feature learning and mapping an input image in a feed-forward manner to its pixel-wise semantic labels. A highlight of the proposed architecture is its ability to produce smooth segment labels when compared with local patch based classifiers. This is due to deep layers of feature encoding that employ a large spatial context for pixel-wise labelling. To the best of our knowledge this is the first deep learning method to learn to map low resolution encoder feature maps to semantic labels. Both qualitative and numerical accuracy of the SegNet for outdoor and indoor scenes is very competitive, even without use of any CRF post-processing. We have also demonstrated the use of pre-trained SegNet for obtaining good performance on other datasets with a small extra computational effort. The encoder-decoder architecture of the SegNet can also be trained unsupervised and to handle missing data in the input during test time.
{\small
\bibliographystyle{ieee}
\bibliography{RefBase}
}

\end{document}